\documentclass[journal,twoside,web]{ieeecolor}
\usepackage{generic}
\usepackage{cite}
\usepackage{amsmath,amssymb,amsfonts}
\usepackage{algorithmic}
\usepackage{graphicx}
\usepackage{algorithm,algorithmic}
\usepackage{hyperref}
\hypersetup{hidelinks=true}
\usepackage{textcomp}

\usepackage{xcolor}
\usepackage{mathtools}
\usepackage{amsmath,bm}
\usepackage[normalem]{ulem}
\usepackage{multirow}

\definecolor{color_C0}{RGB}{31, 119, 180}
\definecolor{color_C1}{RGB}{255, 127, 14}
\definecolor{color_C2}{RGB}{44, 160, 44}
\definecolor{color_C3}{RGB}{214, 39, 40}
\definecolor{color_C4}{RGB}{148, 103, 189}
\definecolor{color_C5}{RGB}{140, 86, 75}
\definecolor{color_C6}{RGB}{227, 119, 194}

\def\BibTeX{{\rm B\kern-.05em{\sc i\kern-.025em b}\kern-.08em
    T\kern-.1667em\lower.7ex\hbox{E}\kern-.125emX}}
\begin{document}
\title{Timing-Aware and Interpretable Self-Attention for Event-Based Wearable Task Transition Detection}
\author{Anda Ouyang, \IEEEmembership{Student Member, IEEE}, Siyuan Chen, \IEEEmembership{Member, IEEE} and Julien Epps, \IEEEmembership{Senior Member, IEEE}
\thanks{The authors are with the School of Electrical Engineering and Telecommunications, University of New South Wales, Kensington, NSW 2033, Australia
(e-mail:\{a.ouyang, siyuan.chen, j.epps\}@unsw.edu.au;)}}

\maketitle

\begin{abstract}

Task transition detection aims to identify when a person shifts from one task to another, providing a foundation for continuous health monitoring and timely context-aware interventions. While statistical and deep feature representations have demonstrated strong classification performance, they often rely on aggregated or latent features that are difficult to interpret in behavioural terms. Event-based features provide a compact and interpretable alternative by encoding salient behavioural and physiological patterns as discrete tokens with explicit semantic meaning, making them well suited for behavioural analysis in health informatics. Self-attention offers a natural mechanism for modelling interactions among such tokens, however conventional self-attention classifiers primarily rely on token order and do not explicitly incorporate real-valued timing or duration information. As a result, temporally informative cues present in event representations may not be fully exploited for task transition detection.
This study proposes an event-based time-aware self-attention framework that represents behavioural signals as sequences of discrete events augmented with onset timestamps and durations. Event timing is encoded using a parameter-efficient kernel-based representation, while a novel type-aware duration embedding modulates information propagation according to event persistence. Experiments on a semi-naturalistic wearable dataset show that the proposed model outperforms existing event-based approaches, achieves performance beyond or comparable with statistical and deep-learning models with a more compact input representation, and provides improved interpretability through integrated-gradient-based analysis. These results indicate that explicit modelling of event timing and duration enhances both classification performance and interpretability, supporting reliable deployment in biomedical and health informatics applications.

\end{abstract}


\section{Introduction}
\label{sec:introduction}
\IEEEPARstart{H}{uman} task transition refers to the shift of attention and behaviour from one task to another and is a fundamental component of daily human activity. Accurate detection of task transitions is a key enabler for many biomedical and health informatics applications, particularly in continuous monitoring and context-aware assistive intervention systems. 
In elderly care, for example, reminders of the change of presbyopic eyeglasses in the transition between sedentary and physical tasks can help reduce the risk of accidental falls \cite{raychoudhury2022activisee}. 
Furthermore, many existing task classification systems operate at the task level \cite{attal2015physical, jiang2015human}, requiring tasks to be manually segmented during experiments---an approach that demands substantial human effort and is impractical for real-life systems. Automatic task segmentation based on transition point identification is not only helpful in reducing annotation effort in research studies, but essential for real-world task classification systems that must function without predefined task boundaries \cite{alhammad2021dynamic}. By identifying transitions between functional tasks, such systems enable quantitative analysis of patient activity patterns, therapy engagement, and recovery progression without manual labeling \cite{chang2020automatic, nguyen2017auto}.

Fixed sensors and wearable sensors are two main approaches to collect physiological and behavioural signals to detect task transition points. While fixed sensors like cameras and infrared motion sensors installed in the environment \cite{feuz2014automated} provide better infrastructure integration in smart environments, e.g. smart house and car drive systems, wearable sensor devices, such as mobile phones \cite{reyes2016transition, saha2021detailed} and smart eyeglasses \cite{raychoudhury2022activisee, ouyang2024evaluation} do not restrict the spatial activity range of users and are therefore more suitable for continuous daily-life monitoring. Among the physiological and behavioral signals collected from wearable sensors, body movement characteristics, for example linear acceleration and angular velocities collected from accelerometers and gyroscopes, and eye characteristics, for example blinking, saccade, and pupil diameter change collected from near-field cameras embedded into smart eyeglasses, were shown to be effective in task transition detection \cite{raychoudhury2022activisee, ouyang2024evaluation}, as they reflect both physical activity execution and cognitive workload.

Although conventional task transition detection systems primarily rely on statistical features \cite{reyes2016transition, raychoudhury2022activisee, saha2021detailed} or deep features \cite{wang2020wearable, khan2025transition}, event-based features, which tokenize continuous-valued signals into discrete events, offer a representation that aligns more closely with how humans perceive and annotate task-related behaviours. Previous studies have shown that event-based features can achieve performance comparable to statistical features while providing enhanced interpretability \cite{ouyang2024evaluation}.
 Moreover, the discrete nature of event-based features enables the application of various Natural Language Processing (NLP) techniques. While the effectiveness of combining Bag-of-Words with a conventional Support Vector Machine (SVM) has been validated \cite{ouyang2024evaluation}, the potential of more advanced NLP models, such as neural network-based architectures, remains underexplored in the context of event-based task transition detection. In parallel, there is a growing trend across domains such as weather forecasting, electrical grid analysis, and traffic modeling \cite{talukder2024totem, liu2024itransformer} to tokenize continuous time series into discrete deep-learned representations---not primarily for interpretability, but to enable compatibility with transformer-based architectures and to facilitate large-scale pre-training and cross-field transfer learning. These approaches have demonstrated consistent performance improvements in imputation, forecasting, anomaly detection and classification.

Among modern NLP techniques, self-attention-based models such as Transformers \cite{vaswani2017attention} and their variants have become dominant in recent research due to their ability to model token-level interactions and their compatibility with parallel computation on GPUs. However, event timing information, which has been demonstrated to be important for task transition detection \cite{ouyang2024evaluation}, is typically absent from conventional NLP token representations. While several recommender systems in e-commerce have proposed mechanisms to incorporate event timestamps into self-attention-based models \cite{xu2019self, li2020time, tran2023attention, cho2020meantime}, such time-aware mechanisms have not yet been explored in the context of event-based task transition detection. Moreover, these models only encode timestamps and do not account for the duration of each event, which may also carry essential temporal cues.


Interpretability is critical for deploying wearable-based machine-learning systems in healthcare and human-centred bioinformatics, where model transparency supports clinical trust, system validation, and principled feature design. For task transition detection in healthcare and health informatics applications, understanding event-level contributions provides insight into how behavioural and physiological changes are used by the model, supporting transparent analysis of task transitions that may reflect changes in cognitive or physical state. Network analysis techniques such as Integrated Gradients (IG) \cite{sundararajan2017axiomatic, hao2021self} enable the analysis of attention mechanisms to reveal token-level contributions, providing a principled basis for interpreting model decisions at the level of individual events. However, such attribution analyses have received limited attention in event-based task transition detection, particularly in conjunction with self-attention models designed to preserve temporal structure.

In this paper, we propose an Event-based Time-Aware Self-Attention model (EvtTiSA) for task transition detection. In addition to adopting the Mercer Kernel approach \cite{xu2019self} to encode event timestamps, we introduce a novel encoding scheme for representing event durations within the self-attention framework. Furthermore, we apply the IG technique to analyse the contribution of individual input events to the model's predictions, enhancing the transparency (interpretability) and reliability of event-based task transition detection in real-world scenarios.

\section{BACKGROUND AND RELATED WORK}
\subsection{Task Transition Detection}

Wearable-sensor-based task transition detection generally comprises data acquisition, feature extraction, and the deployment of machine learning algorithms. For data acquisition, accelerometers and gyroscopes can capture body movement signals such as linear and angular acceleration and velocity across different directions. These signals have been shown to effectively classify non-sedentary physical tasks and detect transitions between them \cite{raychoudhury2022activisee, alhammad2021dynamic, reyes2016transition, ouyang2024evaluation}. However, a large proportion of daily human activities are sedentary in nature, involving limited physical movement while requiring sustained cognitive engagement and frequent task switching. Such sedentary cognitive activities are common in modern lifestyles and are particularly relevant in health-related monitoring scenarios, where prolonged sedentary behaviour and cognitive workload have been associated with fatigue, reduced well-being, and long-term health risks~\cite{ellingson2018changes, saunders2020sedentary}. 

For sedentary tasks, electroencephalography (EEG) can provide a direct neurophysiological measure of task activity \cite{cutini2021time}, however the application and maintenance of electrodes are relatively intrusive and less practical in daily settings. In contrast, eye-related features, collected non-invasively via near-field infrared cameras mounted on smart glasses \cite{raychoudhury2022activisee, ouyang2024evaluation}, can provide indirect indicators of cognitive effort and attention. These systems typically process infrared video from eye cameras to extract indicative features such as pupil diameter, pupil position, blink rate, and their temporal dynamics.

Most existing task transition detection systems adopt either statistical features or end-to-end deep features. Conventional machine learning models, such as Support Vector Machine \cite{reyes2016transition, raychoudhury2022activisee}, k-Nearest-Neighbor \cite{saha2021detailed, raychoudhury2022activisee} and Decision Tree \cite{raychoudhury2022activisee} rely heavily on statistical features, including but not limited to mean, standard deviation, skewness, and energies in different frequency bands. With the prevalence of neural networks, recent research is shifting the focus towards end-to-end deep features due to their demonstrated performance advantages \cite{wang2020wearable, khan2025transition}. For example, Wang et al. \cite{wang2020wearable} employed a CNN model to process body movement signals into deep features map, which were then fed into an LSTM to classify activity types and their transitions. Khan et al. \cite{khan2025transition} arranged their CNN and LSTM modules in parallel and constructed an ensemble by stacking the outputs of the last neural layer. However, deep feature representations are often suffer from limited interpretability and typically require a relatively large amounts of training data.


Event-based features, by contrast, offer improved interpretability compared to both statistical features and deep feature representations \cite{ouyang2024evaluation, chen2020multimodal}. Event-based features are obtained by tokenizing continuous-valued signals, such as angular velocities collected from gyroscopes or pupil diameters extracted from eye videos, into discrete labels. This representation enables compact longitudinal recording of behaviour, since hours of wearable data can be stored and analysed as sparse event sequences in contrast to fully processing every frame of signal data, which is particularly advantageous for resource-constrained wearable systems. Some eye-related features, including blinking, saccade and fixation, naturally occur as events \cite{chen2019task}. 
For body movement, Chen and Epps \cite{chen2019head_event} proposed three atomic head movement events, arguing that head motion is particularly informative for detecting sedentary task processing, where other body parts such as legs, hips, and chest tend to be less active. They derived three head events by determining whether the head is moving and whether the motion is accelerating or decelerating. Ouyang et al. \cite{ouyang2024evaluation} adopted the aforementioned eye and head features into task transition detection, and further proposed features to capture significant pupil size variations within predefined temporal windows. The effectiveness of events and their timing-related features - such as duration and/or inter-event intervals - has been demonstrated to achieve comparable performance to statistical features in various human task processing problems, including task load estimation \cite{chen2019head_event, chen2019task} and task transition detection \cite{ouyang2024evaluation}. Furthermore, human-understandable events facilitate visualisations and more intuitive analysis comparing behavioural patterns during task transition and non-transition periods. For example, saccades tend to become more frequent but shorter in duration near transition points, whereas head movement often remains relatively static for extended periods during non-transition segments \cite{ouyang2024evaluation}. However, the only existing work on event-based task transition detection relies on a conventional SVM model \cite{ouyang2024evaluation}. The application of more advanced NLP architectures and the analysis of individual event contributions within such models remain unexplored.


\subsection{Self-Attention Classifiers}

Since the introduction of the high-performance transformer model \cite{vaswani2017attention}, self-attention-based classifiers have become a dominant model for token sequence processing, including NLP. Although the key advantage of self-attention-based models lies in their ability to efficiently and effectively capture token interactions in large-scale unsupervised pre-training---enabling easy adaptation through transfer learning---they have also demonstrated effectiveness on small datasets without pre-training in various fields \cite{ding2024eeg, guo2020multi}. In classification tasks, only the encoder component of the Transformer is typically employed, as it sufficiently models contextual relationships among input tokens \cite{xu2019self, hao2021self, devlin2019bert}. In this subsection, we review the core architectural components of self-attention models and the standard positional encoding formulations, which will serve as the basis for analysing their timing representations in subsequent sections.


The core component of a transformer is the self-attention block. Given $X = E^{token} \in \mathbb{R}^{N \times d_x}$ representing the input sequence where each row $E_{i,:}^{token}$ contains an embedded token, and $W^Q, W^K, W^V \in \mathbb{R}^{d_x \times d_k}$ as parameter matrices, the self-attention can be formulated as:

\begin{equation}
Q=XW^Q, K=XW^K, V=XW^V
\label{eq_KQV}
\end{equation}

\begin{equation}
A = \text{softmax}\left(S\right)=\text{softmax}\left(\frac{QK^\top}{\sqrt{d_k}}\right)
\label{eq_A}
\end{equation}

\begin{equation}
H=AV
\label{eq_H}
\end{equation}

where the $i$th row $j$th column element of the $A$ matrix, denoted $A_{i,j}$, is the interaction from token $i$ to token $j$ and $A_{j,i}$ is vice versa: $A_{i,j}$ is the inner product of linearly transformed token $i$ and token $j$ after a per-row softmax normalization, further multiplied with the token $j$ in the value matrix in equation (\ref{eq_H}).

The self-attention mechanism described above is inherently order-invariant, meaning that it cannot capture the positional relationships between tokens in a sequence. To address this limitation, positional encodings can be incorporated to inject information about token order into the model. Two widely adopted approaches are absolute positional encoding and relative positional encoding. Absolute positional encodings \cite{vaswani2017attention, devlin2019bert} represent the absolute index of each token as a vector, which is then added to or concatenated with the token embeddings in $X$ in equation~(\ref{eq_KQV}). For the commonly used additive formulation, the input to the self-attention layer becomes

\begin{equation}
X = E^{token} + E^{pos}
\label{eq_X_absolute_position}
\end{equation}

Relative positional encodings \cite{shaw2018self} instead embed the position difference between tokens into $M \in \mathbb{R}^{N \times N \times d_x}$ such that $M_{i,j,:}=\mathrm{encode}(|i-j|)$, and reformat equations (\ref{eq_A}) and (\ref{eq_H}) into:

\begin{equation}
A = \text{softmax}\left(\frac{QK^\top + Q \bullet M}{\sqrt{d_k}}\right)
\label{eq_A_relative}
\end{equation}

\begin{equation}
H=AV+A \bullet M
\label{eq_H_relative}
\end{equation}


where $\bullet$ denotes a last-dimension dot product.
Specifically, $(Q \bullet M)_{i,j} = Q_{i,:} M_{i,j,:}^\top$, and $(A \bullet M)_{i,j} = A_{i,:} M_{i,j,:}^\top$.


With the absolute or relative positional encoding, $A_{i,j}$ becomes a learned function of both token content and positional information, rather than a pure similarity measure in embedding space. This positional dependency fundamentally alters the structure of the attention matrix, enabling the model to encode order-sensitive and distance-aware interaction patterns that cannot be represented by content-only self-attention.

\subsection{Engaging Timing Information}

Standard attention-based NLP models primarily encode the sequential order of tokens and do not explicitly account for real-valued timing information, which could be useful for modelling task transition. In natural language, tokens are treated as discrete symbolic units without associated physical timestamps; consequently, modelling their positional order is typically sufficient for capturing syntactic and semantic dependencies. Nevertheless, timing information has been explicitly incorporated in temporal recommender systems, such as those for movies, music, and e-commerce~\cite{xu2019self, li2020time, tran2023attention, cho2020meantime}. In these application contexts, user interactions---such as ratings, clicks, or purchases---occur at irregular and heterogeneous time intervals, and modelling such temporal dynamics has been shown to improve recommendation performance.


Existing methods for incorporating timing information are derived from the absolute and/or relative order positional encoding. Xu et. al. \cite{xu2019self} proposed a parameter-efficient Mercer kernel to embed arbitrary absolute timestamps into a Fourier cosine map, and Tran et. al. \cite{tran2023attention} further fused this formulation with an LSTM to capture long-term dependencies of events. Li et. al. \cite{li2020time} proposed a relative timing encoding scheme (TiSASRec) by replacing the order position differences with quantized inter-event time intervals in Eq.~(\ref{eq_A_relative}), such that $M_{i,j,:}=\mathrm{LinearEmbed}(\mathrm{quantize}(|t_i-t_j|))$. Building upon this idea, Cho et. al. (MEANTIME) \cite{cho2020meantime} replaced the linear embedding with non-linear sinusoidal, logarithmic, and exponential kernels to model non-linear temporal interactions. Incorporating timing information in these models has been shown to improve recommendation performance in an e-commerce context, as measured by hit rate, to different degrees. 

However, to the best of the authors' knowledge, none of these timing-aware self-attention methods have been applied to human task processing problems, including task transition detection. Moreover, because events in recommender systems are typically represented as instantaneous timestamped interactions, \textit{none of the aforementioned self-attention methods account for the event duration}. In contrast, event duration has been shown to provide discriminative information in human task processing~\cite{ouyang2024evaluation}. However, how to incorporate event duration into self-attention classifiers remains an open problem.

\subsection{Analysis of Attention via Attribution Methods}
Attribution analysis aims to quantify how input features or intermediate representations contribute to a model’s output. Early approaches such as gradient-based saliency maps~\cite{simonyan14a}, Layer-wise Relevance Propagation (LRP)~\cite{binder2016layer}, and Grad-CAM~\cite{selvaraju2017grad} primarily provide relative importance patterns and lack a well-defined absolute scale. Consequently, their attribution values are most meaningful when compared across features or layers, rather than as direct measures of contribution to the model output.

Integrated Gradients (IG)~\cite{sundararajan2017axiomatic} address this limitation by introducing an attribution framework grounded in a completeness property. For a system $y = F(\bm{x})$ with input or intermediate variable $\bm{x}$ and a reference point $\bm{x}'$, the integrated gradient is defined as
\begin{equation}
\mathrm{IG}(\bm{x}) := (\bm{x} - \bm{x}') \odot \int_{0}^{1} 
\frac{\partial F\big(\bm{x}' + \alpha (\bm{x} - \bm{x}')\big)}{\partial \bm{x}} 
\, d\alpha 
\label{eq_IG}
\end{equation}

where $\odot$ denotes element-wise (Hadamard) multiplication. The integrated gradient satisfies the completeness property ~\cite{sundararajan2017axiomatic}, such that the sum of $\mathrm{IG}$  across all dimension equals the output difference:

\begin{equation}
\sum_i \mathrm{IG}(\bm{x}_i) = F(\bm{x}) - F(\bm{x}') 
\label{eq_delta_y}
\end{equation}

This property enables each attribution component to be interpreted as a quantifiable contribution to the output change relative to the chosen reference point $\bm{x}'$. 

For Transformer-based models, interpretability is often analysed through attention-weight visualisations \cite{ji2021usevis, vig2019multiscale}. However, several studies have questioned whether raw attention scores reliably reflect feature importance~\cite{jain2019attention, wiegreffe2019attention}. One contributing factor is that attention outputs are subsequently transformed by learned projections and feedforward layers, weakening the direct correspondence between attention magnitude and final prediction.

To address this limitation, Hao et al.~\cite{hao2021self} extended Integrated Gradients to the self-attention mechanism by treating the attention matrix as an intermediate activation and computing gradient-based attributions with respect to it. Unlike raw attention-weight visualisation, this formulation attributes attention elements based on their propagated effect on the final model output, which can be written as:

\begin{equation}
\mathrm{Attr}(A^{i_{\text{head}}}) := A^{i_{\text{head}}} \odot 
\int_{0}^{1} 
\frac{\partial F\big( \alpha A^{i_{\text{head}}}\big)}{\partial A^{i_{\text{head}}}} 
\, d\alpha 
\label{eq_IG_A}
\end{equation}

Their results show that IG-based attention attributions better reflect the functional importance of attention heads than raw attention weights, particularly in the context of multi-head pruning. Despite this advance, existing attention attribution methods implicitly assume that the chosen reference point corresponds to a consistent neutral model state~\cite{sundararajan2017axiomatic}. Specifically, the reference output $F(\bm{x}')$ should represent the absence of contribution, such that positive and negative IG values can be interpreted as contributions toward and against a given prediction, respectively. Moreover, this reference output must be consistent across attribution instances, so that attribution values reflect output changes from a common baseline and are therefore directly comparable across different inputs or model components.

In standard Transformer architectures, however, this assumption is difficult to satisfy, as the model output generally depends on multiple parallel attention heads and residual pathways that remain active under common attribution settings. The architectural origins of this non-neutrality are discussed in detail in Section~III.D. Consequently, the reference output may already reflect a non-neutral prediction state, meaning that the resulting IG attributions quantify changes relative to an already active baseline rather than the absence of contribution. This baseline dependency can introduce offsets in the attribution scores and undermine their interpretability. As a result, the completeness interpretation of IG-based attention attributions may no longer hold in practice, leaving open the question of how to obtain faithful and interpretable attention attributions under realistic architectural settings. Nevertheless, this issue can be addressed through careful architectural design that enforces a well-defined neutral reference state.

\section{METHODS}

\begin{figure*}[t]
\centering
\includegraphics[width=\textwidth]{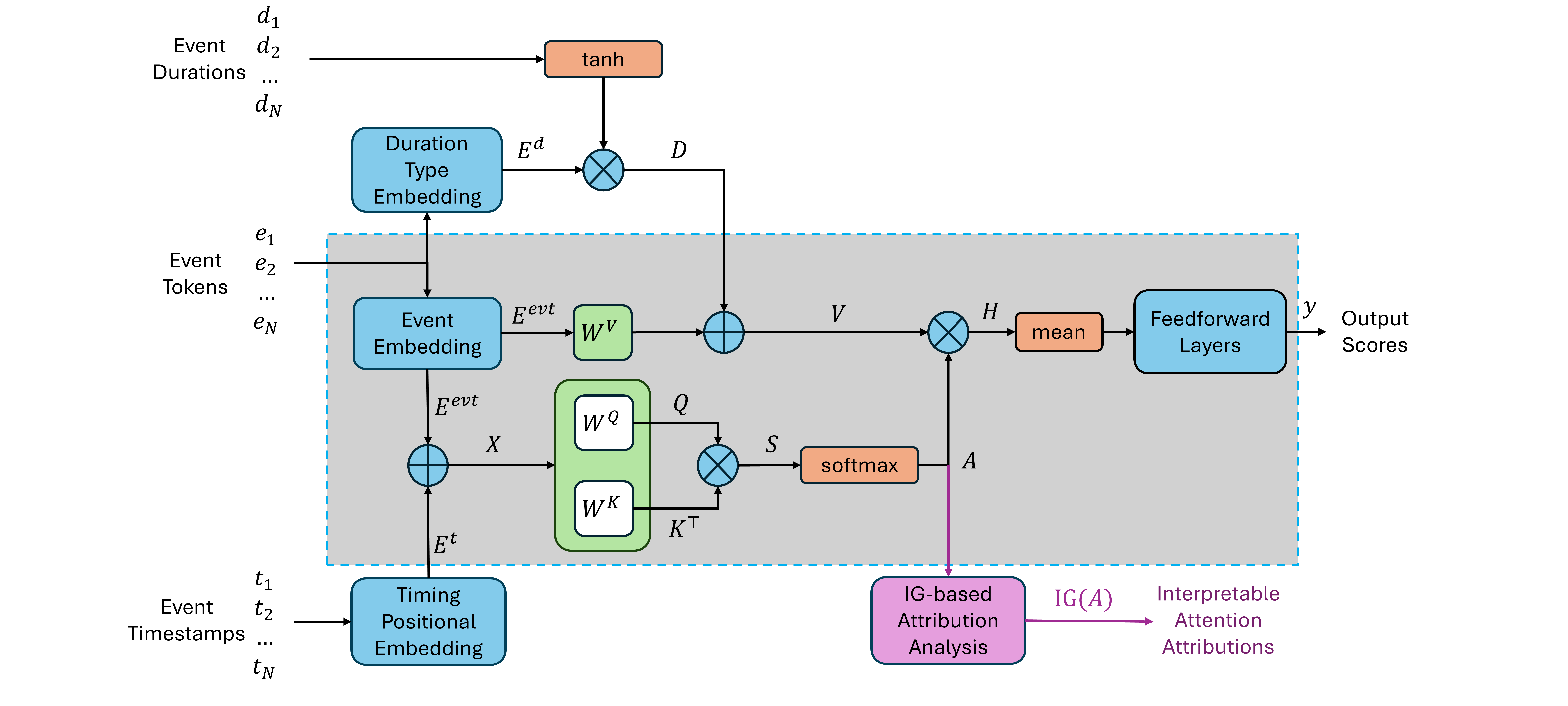}
\caption{Overview of the proposed EvtTiSA system that combines events, event timestamps, and event duration information in attention. The structure inside the light grey box corresponds to a conventional self-attention classifier (with the skip connection from $X$ to $H$ removed to enable integrated-gradient-based attribution analysis). EvtTiSA (1) replaces order positional embeddings with timing positional embeddings to capture temporal interactions between events, (2) incorporates event duration information $D$ into the value matrix $V$, and (3) enables post-hoc IG-based Attribution Analysis for attention mechanism interpretability.}
\label{fig:block_diagram}
\end{figure*}

In this paper, we propose a self-attention-based classifier to combine event tokens, timing, and duration. As shown in Fig. \ref{fig:block_diagram}, 
the grey box area is the structure of a conventional self-attention classifier, from which basis we 1) replace the order position embedding with timing positional embedding in the structure below the grey box, to capture the timing interactions of events; 2) propose an event duration embedding system in the structure above the grey box, to engage event duration information into the classfier. More details and reasoning for the structure design are explained in the following subsections. The inputs to the system are $[e_1,e_2,...,e_N]$, a list of event onset token indices for a sliding window, $[t_1,t_2,...,t_N]$, where each $t_i$ represents the timing position or onset timestamps of $e_i$, and $[d_i,d_2,...,d_N]$, where each $d_i$ represents the timing duration of event $e_i$. The self-attention output is passed through a two-layer feedforward network with ReLU activation, followed by a sigmoid activation to produce the final score $y \in [0, 1]$. A threshold is applied to this score for binary classification. 

\subsection{Signal Tokenization: Event-based Features}

Event-based features aim to represent continuous-valued physiological and behavioural signals as discrete, semantically meaningful events. Such representations not only improve interpretability but also enable the application of a wide range of NLP models. In prior work, event-based features were commonly extracted by converting continuous signals into frame-based binary indicator sequences \cite{chen2019head_event, ouyang2024evaluation}. In this representation, each event type is encoded as a binary channel over a fixed temporal grid with the same temporal resolution as the original signals. While this representation is convenient for event detection and visualisation, modern NLP models typically operate on token sequences rather than multidimensional time-aligned arrays.

Representing behavioural data as a sequence of event tokens enables direct attribution of model outputs to individual events. In contrast, frame-based binary indicator sequences represent a single event across multiple consecutive time frames, causing its influence to be distributed over many input positions and thereby diluting its attribution and reducing interpretability. Another advantage is that event-token sequences reduce input dimensionality because the number of inputs no longer scales with event duration, thereby lowering computational cost and improving statistical efficiency. Although some existing approaches further convert binary indicators into event-onset sequences, they often discard event duration or relative timing information, which are essential for event-driven sequence modelling. To address these limitations, we propose a temporal event representation in which each event is explicitly characterised by its type and timing information.

The proposed temporal event sequence is constructed from event indicator sequences. Following \cite{ouyang2024evaluation}, we extract ten event indicator sequences corresponding to head movement, eye movement, and pupil dynamics:


\textbullet ~$head\_central, head\_increase, head\_decrease$: Head movement events are extracted using the algorithm proposed by Chen and Epps \cite{chen2019head_event}. Let $\omega_h$ denote the angular velocity magnitude of head movement. A threshold of $5^\circ/s$ is applied to distinguish voluntary head movement from involuntary micro-oscillations. Head movement events are defined as

\begin{equation}
e^{h}=\begin{cases}
      head\_central & \text{if $\omega_{h}\leqslant5^{\circ}/s$}\\
      head\_increase & \text{if $\omega_{h}>5^{\circ}/s$, $t^{pc}<t<t^{pk}$}\\
      head\_decrease & \text{if $\omega_{h}>5^{\circ}/s$, $t^{pk}\leqslant t<t^{nc}$}
    \end{cases} 
\end{equation}

where $t^{pc}$ and $t^{nc}$ denote the times of the previous and next $head\_central$ events, and $t^{pk}$ denotes the time of peak angular velocity.

\textbullet ~$is\_blink, not\_blink$: Blinking is detected by the eye-landmark algorithm \cite{chen2023high}.

\textbullet ~$is\_saccade, not\_saccade$: Saccade and fixation are extracted from the dispersion-based algorithm \cite{salvucci2000dispersion}.

\textbullet ~$pupil\_size\_central$, $pupil\_size\_increase$, $pupil\_size$ $\_decrease$: Pupil size changes are detected by identifying diameter variations exceeding 7\% within 2~s, following the principle of threshold-based change detection in \cite{ouyang2024evaluation}, using a relative change criterion.

For visualisation purposes, each event type is denoted by an abbreviation listed in Table~\ref{tab:token-abbreviations}. After event extraction, we obtain 10 frame-based binary-valued event indicator sequences, one for each event type, indicating whether the corresponding event is active at each time sample.

\begin{figure}[pt]
\centerline{\includegraphics[width=\columnwidth]{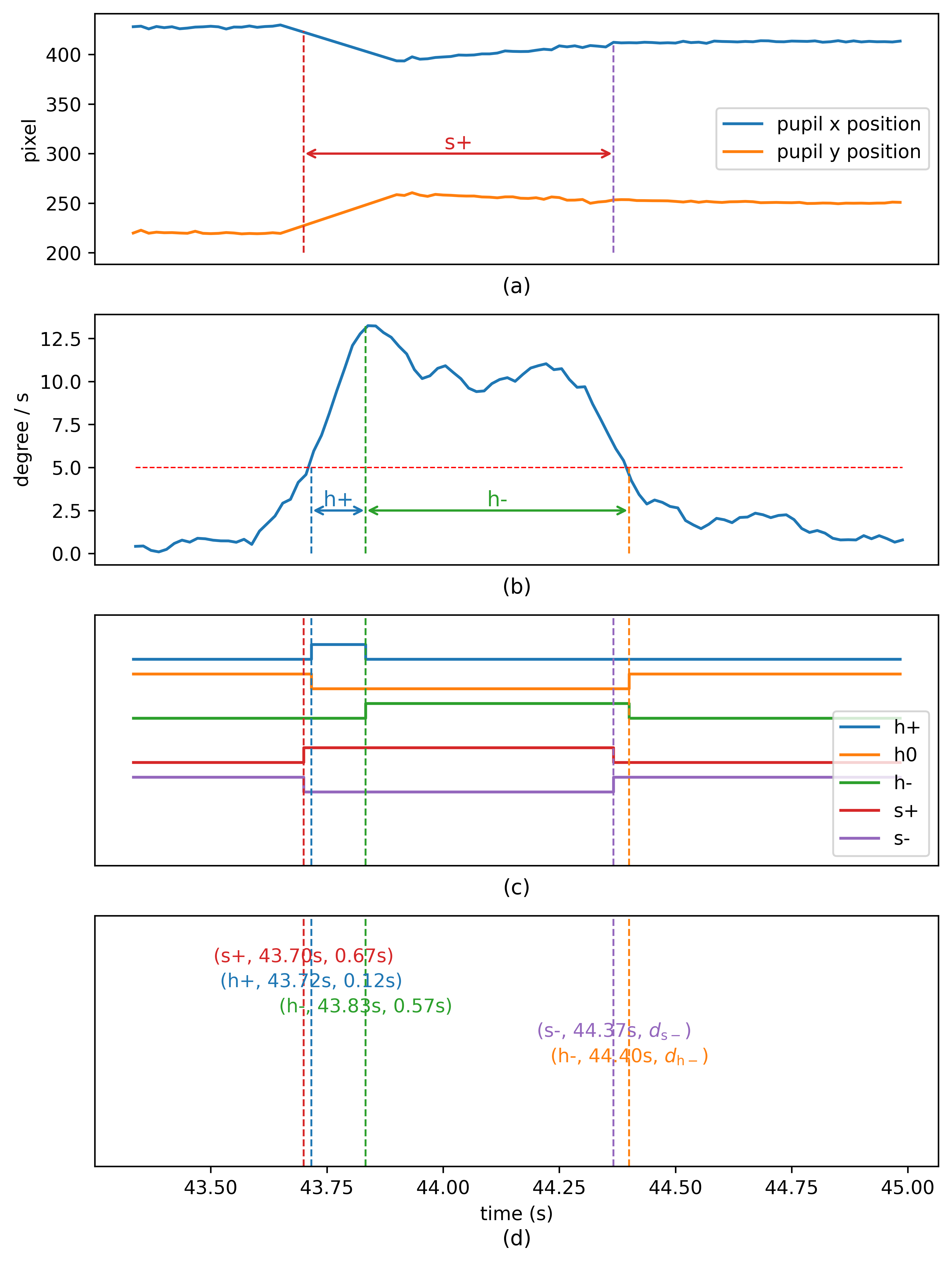}}
\caption{Converting continuous-valued sensor-derived feature streams into temporal events.
(a) Pupil center position time series (pixels). (b) Head angular velocity time series (deg/s).
(c) Uniform-time binary-valued event indicator sequences ($0/1$; false/true for each event type at each time sample), extracted from the continuous signals in (a) and (b); contiguous runs of ones correspond to event instances.
(d) Temporal event sequence obtained by grouping the indicator sequences in (c), serving as the input to the proposed EvtTiSA. Each event instance is represented as a tuple $(e_i, t_i, d_i)$, where $e_i$ is the event type, $t_i$ is the onset timestamp, and $d_i$ is the duration; the model input is the ordered sequence $\{(e_i, t_i, d_i)\}_{i=1}^N$ in ascending temporal order of $t_i$.}
\label{fig:time_series_to_events}
\end{figure}

Fig.~\ref{fig:time_series_to_events} illustrates the conversion from continuous sensor signals to event-based representations.
In (a), the dispersion-based algorithm identifies a saccade within the time interval bounded by the red and purple dashed lines.
The corresponding event indicator sequence in (c) shows the $s+$ ($is\_saccade$) indicator activated during this interval, with its complement $s-$ ($not\_saccade$) inactive.
Similarly, (b) shows a head movement segment in which head acceleration ($h+$) and deceleration ($h-$) phases are identified, whose corresponding indicator sequences are also shown in (c).
These examples visually illustrate how continuous-valued signals are converted into binary-valued event indicator sequences prior to temporal event construction.

Rather than using the binary indicator sequences directly, we group each contiguous active segment into a temporal event instance characterised by its event type, onset time, and duration, as shown in Fig.~\ref{fig:time_series_to_events}(d). Specifically, each event instance is represented as a tuple $(e_i, t_i, d_i)$, where $e_i$ denotes the event type, $t_i$ the onset timestamp, and $d_i$ the event duration. The resulting representation is an ordered sequence of temporal event tuples $\{(e_i, t_i, d_i)\}_{i=1}^N$, sorted by ascending $t_i$, which serves as the input to the proposed EvtTiSA model. This representation aligns with the token-based processing paradigm of self-attention models, facilitates event-level attribution analysis, and explicitly preserves timing information for subsequent encoding. For example, the temporal event sequence in Fig.~\ref{fig:time_series_to_events}(d) is \{$(s+,43.70 s,0.672 s)$, $(h+,43.72 s,0.12 s)$, $(h-,43.83 s,0.57 s)$, $(s-,44.37 s,d_{s-})$, $(h-,44.40 s,d_{h-})$\}.

\begin{table}[htbp]
\caption{Event Token Abbreviations}
\begin{center}
\begin{tabular}{|c|c||c|c|}
\hline
\textbf{Event Type} & \textbf{Abbrev.} & \textbf{Event Type} & \textbf{Abbrev.} \\
\hline
$head\_central$         & $h0$ & $pupil\_size\_central$  & $p0$ \\
$head\_increase$        & $h+$ & $pupil\_size\_increase$ & $p+$ \\
$head\_decrease$        & $h-$ & $pupil\_size\_decrease$ & $p-$ \\
$is\_blink$             & $b+$ & $is\_saccade$           & $s+$ \\
$not\_blink$            & $b-$ & $not\_saccade$          & $s-$ \\
\hline
\end{tabular}
\label{tab:token-abbreviations}
\end{center}
\end{table}

\subsection{Timestamp Embedding: Mercer Kernel}

To embed the non-uniform timing positions or timestamps of events, we adopt the Mercer timing kernel proposed in e-commerce by Xu et al.\cite{xu2019self} for its parameter efficiency, approximate translation invariance, positive semidefiniteness, and demonstrated effectiveness in modelling temporal dependencies. An arbitrary timestamp $t$ is mapped to a vector to match the dimensionality of the event token embeddings:

\begin{equation}
\mathcal{K}(t)=[\sqrt{c_{2i}} \cos(\omega_i t),\sqrt{c_{2i+1}}\sin(\omega_i t)]_{i=1}^{d_x}\label{eq_mercer}
\end{equation}

The Fourier coefficients $c_i$ are learnable parameters that adaptively weight the importance of different temporal frequencies. The frequencies $\omega_i$ can be fixed as hyperparameters or can be jointly optimized with the network parameters; in either case they need to be initialized across a wide range (e.g. logarithmic scale \cite{xu2019self}) to capture interactions across multiple temporal scales. This practice can be understood in light of the highly non-convex nature of frequency learning, in which multiple local optima can arise. As an example, for periodic signals, frequencies at integer multiples of a base frequency (harmonics) can act as local minima, causing optimization to stall near them.

Given a sequence of event timestamps $\bm{t}=[t_1,t_2,\ldots,t_N]$, the Mercer kernel produces a timestamp embedding matrix

\begin{equation}
E^t = \mathcal{K}(\bm{t})
\label{eq_Et}
\end{equation}

The timestamp embedding $E^t$ is added to the embedded event onset tokens $E^{evt}$ to form the query and key matrices, so that the resulting attention scores will capture the interactions of both event type and event timing:

\begin{equation}
Q=(E^{evt}+E^t)W^Q, K=(E^{evt}+E^t)W^K
\label{eq_qk_t}
\end{equation}

\begin{equation}
S_{i,j} = \frac{Q_{i,:}K_{j,:}^\top}{\sqrt{d_k}}
=\frac{(E_{i,:}^{evt}+E_{i,:}^t)W^Q((E_{j,:}^{evt}+E_{j,:}^t)W^K)^\top}{\sqrt{d_k}}
\label{eq_s_t}
\end{equation}

\subsection{Duration Embedding}

The effectiveness of event durations has been demonstrated in \cite{ouyang2024evaluation}. However, to the best of the authors' knowledge, explicit duration embeddings have not been explored in self-attention-based sequence models. Existing timing-aware self-attention architectures have primarily been developed in the context of recommender systems \cite{xu2019self, li2020time, cho2020meantime}, where event sequences (e.g., shopping or website click logs) are typically characterised by timestamps but lack well-defined event durations.

Rather than simply adding or concatenating the event durations to some layers of the system, we propose a type-aware duration embedding scheme, in which duration information is modulated according to event type. As shown by the section above the grey block in Fig. (\ref{fig:block_diagram}), each event onset token is first embedded into a duration-type-specific representation $E_{i,:}^d$, which is then scaled by a nonlinear function of the event duration $d_i$. We employ a hyperbolic tangent function to introduce  nonlinearity and prevent long-duration events from overwhelming the short-duration ones:

\begin{equation}
D_{i,:}=\tanh(\alpha d_i)E_{i,:}^d
\label{eq_D}
\end{equation}

where $\alpha$ is a learnable scaling parameter. The resulting duration embedding matrix $D$ is added to the value matrix $V$:

\begin{equation}
V=E^{evt}W^V+D
\label{eq_V_dura}
\end{equation}

so that event duration influences the amount of information propagated through attention, while the attention weights themselves remain determined by event semantics and timing.

\subsection{Attention Attribution with a Neutral Reference}

Although IG-based attention attribution provides a principled alternative to raw attention analysis, its faithfulness depends on the choice of reference point. For the completeness property of Integrated Gradients to admit a meaningful interpretation, the reference output should correspond to a neutral and consistent model state, so that attribution values represent contributions relative to the absence of influence and identical attribution magnitudes correspond to identical output shifts. In Eq.~(\ref{eq_IG_A}), Hao et al.~\cite{hao2021self} implicitly adopt $A'=\bm{0}$ as the reference attention matrix by scaling one attention head at a time while keeping all remaining heads unchanged. However, in standard Transformer architectures this choice does not correspond to a consistent neutral model state, thereby undermining the completeness interpretation of the resulting attributions.

The non-neutrality of the reference point $A'=\bm{0}$ arises from two architectural factors. First, in multi-head attention, masking or scaling a single head while leaving the remaining heads active produces a reference output that still contains non-zero attention contributions. Consequently, the reference output $F(A'=\bm{0})$ varies across heads, and the completeness property no longer holds consistently. Second, many self-attention-based models incorporate residual or skip connections that bypass the attention module, allowing the input token representations $\bm{X}$ to directly influence downstream feedforward layers~\cite{vaswani2017attention, xu2019self, shaw2018self, li2020time}. Even when $A=\bm{0}$, these bypass pathways can contribute substantially to the final output, further violating the assumption that $F(\bm{0})$ represents a neutral baseline.

To ensure a well-defined neutral reference for attention attribution, we impose two architectural constraints. First, we adopt a single-head attention mechanism, eliminating inter-head interactions and ensuring that all attention contributions are simultaneously removed when $A=\bm{0}$. Second, we remove the skip connection from the input $\bm{X}$ to the post-attention representation $\bm{H}$ (Fig.~\ref{fig:block_diagram}), such that the model output depends solely on the attention-mediated pathway. Under these conditions, setting $A=\bm{0}$ yields a fixed neutral reference output $F(\bm{0})$, and the integrated gradients computed in Eq.~(\ref{eq_IG_A}) faithfully quantify the contribution of each attention element to the final prediction.

\section{EXPERIMENTAL SETUP}

\subsection{Dataset and Pre-processing}

Experiments were conducted on a semi-naturalistic dataset in which the everyday activities of 11 participants were recorded in a laboratory environment for durations ranging from 38 to 89 minutes (M = 66.1, SD = 17.8) \cite{ouyang2024evaluation}. Informed consent was signed by each participant prior to the experiment and all the processes were approved by the UNSW Human Research Ethics Committee (approval date: 9 September 2014; HC No. 08/2014/23). To enable data collection without constraining participants’ behaviour, an unobtrusive eyeglasses-based device, the Pupil Labs Pupil Core \cite{kassner2014pupil}, was employed. The device integrates near-field infrared cameras for eye video capture, along with a triaxial accelerometer and gyroscope for recording head movements. Compared with tightly controlled experimental task designs \cite{wang2020wearable, khan2025transition}, this dataset captures a broader range of spontaneous behaviours that more closely reflect natural activity patterns. The most frequent activities during the recordings were sedentary and included, but were not limited to, laptop use, mobile phone use, reading, and writing. In total, the dataset comprises 12 hours and 7 minutes of recordings, with 725 manually annotated task transitions as ground truth.

To prepare the multimodal data for analysis, time-series features were extracted from the eye videos and inertial signals were preprocessed for temporal alignment. An eye-landmark detection algorithm \cite{chen2023high} was used to derive pupil diameter, pupil centre positions, and blink events from the raw eye videos. Linear and angular accelerations and velocities from the accelerometer and gyroscope were resampled to 60~Hz to match the sampling rate of the eye-derived features. The event detection algorithms described in Section~III-A were then applied to convert the continuous signals into sequences of discrete events.

\subsection{Task Formulation and Evaluation Protocol}

Task transition detection was formulated as a binary classification problem using a sliding-window approach, where each window was labelled according to whether a task transition occurred within it. Following \cite{ouyang2024evaluation}, a window length of 2~seconds and a step size of 1~second were adopted. To alleviate class imbalance, transition windows were oversampled by reducing the step size to 4 frames (66.67~ms), resulting in transition samples accounting for 33.7\% of the total 63{,}113 sliding windows. Performance was evaluated using 11-fold leave-one-participant-out cross-validation.

To evaluate the effectiveness of timestamp embedding, we compared the Mercer timing kernel \cite{xu2019self} in Eq.~(\ref{eq_mercer}) with the sinusoidal positional encoding \cite{vaswani2017attention} for representing temporal order and timing information. The embedding dimension was swept from 20 to 120 in increments of 20 to identify a suitable dimensionality. To reduce the influence of random initialisation, each configuration was trained 15 times with different random seeds, and the average performance was reported.

As for the analysis of event attribution, integrated gradients were computed for each trained model and averaged across the 165 models obtained from 11-fold leave-one-participant-out cross-validation with 15 random seeds per fold. The consistency and reproducibility of attribution values across models were quantified using the one-way random-effects intraclass correlation coefficient (ICC)~\cite{koo2016guideline}. Specifically, ICC(1,1) was used to assess the reliability of attribution values from a single trained model instance, while ICC(1,k) was used to evaluate the reliability of the averaged attribution across all 165 model instances. This formulation is appropriate as each trained model can be regarded as a random realisation of the same training procedure. In addition, attribution-based feature pruning experiments were conducted by retaining either the most contributing or the least contributing features, in order to evaluate whether the attribution scores reflect meaningful feature importance.

The Area Under the Receiver Operating Characteristic Curve (AUC) was adopted as the primary evaluation metric, as it is independent of decision threshold selection. Accuracy and F1 score were additionally reported to provide complementary perspectives on classification performance. Reported results correspond to the mean performance across the 11 leave-one-participant-out folds, with standard deviations reflecting inter-participant variability.

Paired Wilcoxon signed-rank tests were conducted on the fold-wise AUCs to assess whether performance differences between EvtTiSA and reference models were statistically significant.

\subsection{Model Configuration and Implementation}

Regarding hyperparameter settings, the angular frequencies $w_i$ for both the Mercer kernel and the sinusoidal encoding were fixed on a logarithmic scale from $1 \times 10^{-4}$ to $1$, following \cite{vaswani2017attention}, enabling the models to capture interactions across both short and long temporal scales. All Fourier coefficients $c_i$ in the Mercer timestamp embedding formulation in Eq.~(\ref{eq_mercer}) were initialised to 1. The duration scaling parameter $\alpha$ in Eq.~(\ref{eq_D}) was initialised to 0.1 so that the duration embedding does not saturate for durations shorter than approximately 15~seconds at initialisation. All remaining parameters were randomly initialised, with their exact values determined by the random seed.

All models were trained on NVIDIA V100 GPUs on the Gadi supercomputing cluster at the National Computational Infrastructure (NCI), Australia. The Adam optimiser was employed due to its robustness and stable convergence. As 99\% of the sliding windows contained fewer than 30 events, event sequences were truncated to a maximum length of 30 and zero-padded where necessary to form fixed-length batches. To exclude padded tokens from attention computation, negative infinity was applied to the corresponding entries in the score matrix $S$, and zero values were applied in the attention matrix $A$, i.e., $S_{i,j} = -1 \times 10^{9}$ and $A_{i,j} = 0$ whenever either $e_i$ or $e_j$ corresponded to a padded token. Early stopping was applied with a tolerance of 0.5\% loss change over 10 consecutive epochs, with training typically converging within 40--80 epochs.



\section{RESULTS AND DISCUSSION}
\begin{figure}[pt]
\centerline{\includegraphics[width=\columnwidth]{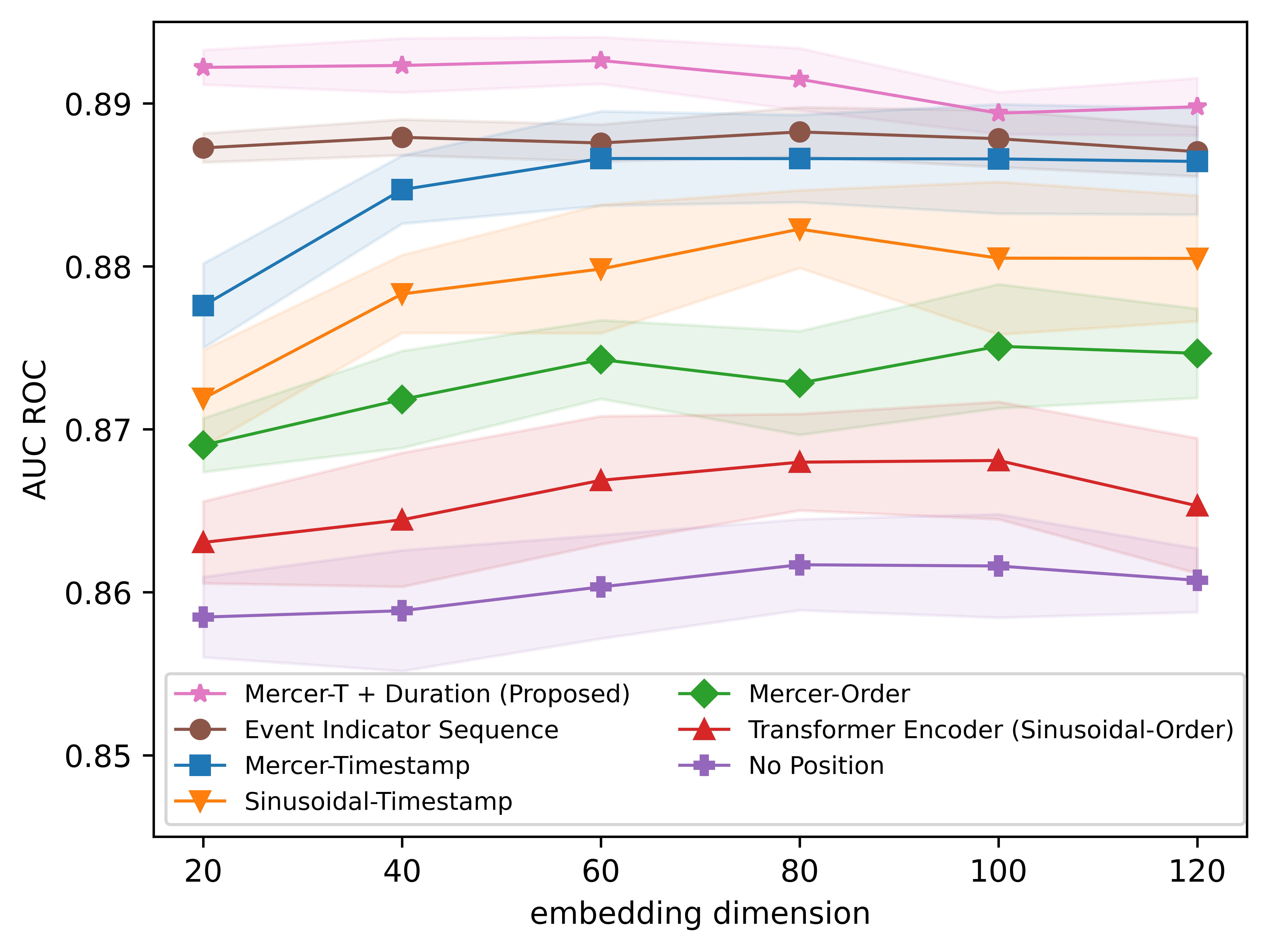}}
\caption{AUC performance comparison of Mercer and sinusoidal encoding for timing and order information, and no position embedding baseline, with the variation of embedding dimensions. The solid lines are the average performances across 15 different random seeds, and the light colour regions are $\pm$ standard deviation. All models, except for the "event indicator seq" (uniform sampled event indicator sequence), use only event onsets occurring at irregular (non-uniform) timestamps. In contrast, the uniform timestep model uses the full event sequence sampled at a fixed rate of 60 Hz. } 
\label{fig:auc_pos}
\end{figure}

\subsection{Comparison of Order Embeddings and Timing Embeddings}

\begin{figure*}[pt]
\centerline{\includegraphics[width=\textwidth]{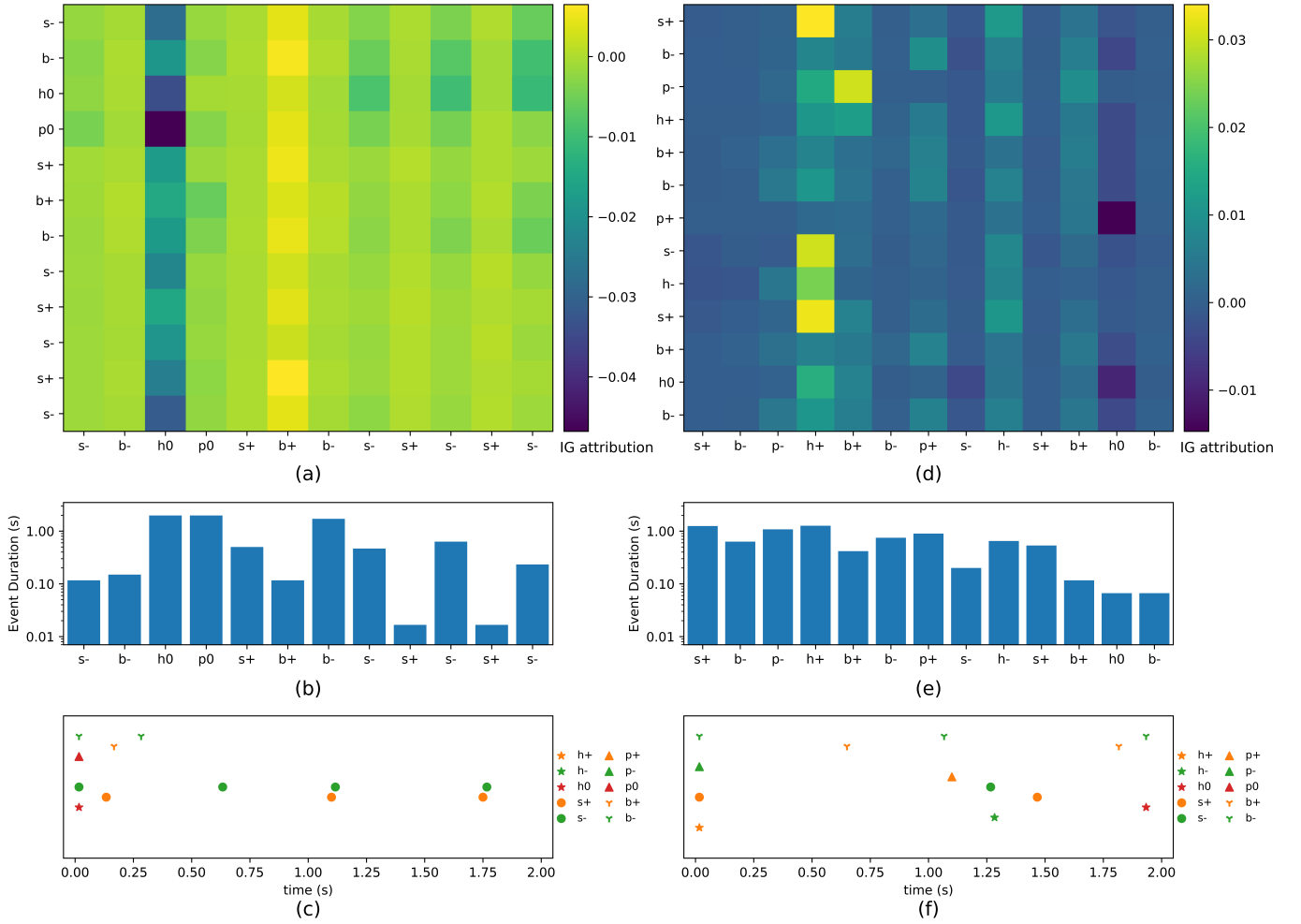}}
\caption{Attribution insights, where a) is the integrated-gradient-based attribution heatmap of a \textit{non-transition} sliding window, b) is the duration of each event onset of the same sliding window, and c) is the timestamps of each event onset of the same sliding window. d) e) f) \textit{}are those of a transition sliding window}  
\label{fig:ig_both}
\end{figure*}

We conducted an ablation study of the proposed EvtTiSA model to examine how order-based and timing-based position and its embedding dimensions impact performance, by comparing sinusoidal embeddings and Mercer kernel embeddings across different embedding dimensions. As shown in Fig.~(\ref{fig:auc_pos}), the performance of all embedding schemes becomes stable for embedding dimensions greater than or equal to 40, indicating that these dimensions are sufficient to capture the information contained in event types together with their temporal or ordinal positions.

From the {\color{color_C0}$\blacksquare$}, {\color{color_C1}$\blacktriangledown$}, {\color{color_C2}$\blacklozenge$} and {\color{color_C3}$\blacktriangle$} curves, timing embeddings consistently outperformed order embeddings under both encoding mechanisms, while the Mercer kernel consistently outperformed the sinusoidal embedding for both timing- and order-based representations across embedding dimensions. Although the resulting performance gains are modest, both effects are systematically observed, suggesting that the absolute timing of event onsets captures marginally richer task transition–related information than relative ordering alone, and that the Mercer kernel provides a more flexible temporal representation.

Inspection of the learned Mercer kernel coefficients reveals a smooth, scale-dependent structure that consistently emerges across cross-validation folds and random initialisations, with slightly greater emphasis on mid-scale temporal variations (approximately 0.3–2 s) and modest attenuation of higher-frequency variations (approximately 0.02–0.2 s). This suggests that discriminative information in the dataset is not concentrated in instantaneous micro-fluctuations, but rather in short-range temporal grouping of events spanning sub-second to few-second intervals. The improvement over sinusoidal encoding therefore likely arises from the ability to adaptively suppress very high-frequency variations while modestly emphasising intermediate temporal dependencies.

To further examine whether temporal event tuples (Fig.~\ref{fig:time_series_to_events}(d)) can sufficiently capture the information contained in uniformly sampled event indicator sequences (Fig.~\ref{fig:time_series_to_events}(c)), we compared in Fig.~\ref{fig:auc_pos} the performance of the uniform time-spaced event indicator sequence model ({\color{color_C5}$\bullet$}) with models that rely solely on event onsets and their timestamps or orders. We encoded the uniform timestamps of the event indicator sequence using the Mercer kernel to ensure consistency with the proposed timing representations and to enable a fair comparison. When compared with the event-onset-only model using the same Mercer timestamp encoding ({\color{color_C0}$\blacksquare$}), the AUC decreased by only 0.005.

By further incorporating event duration information through the proposed complete EvtTiSA ({\color{color_C6}$\star$}) formulation in equations (15) and (16), the average AUC increased to 0.909 for embedding dimensions 20–80, slightly outperforming the event indicator sequence model. Importantly, this temporal event representation reduces the input size from a $10 \times 120$ event indicator sequence to a compact $3 \times 30$ tuple sequence per sliding window. Given the limited computational resources and memory capacity of wearable devices, this lightweight representation improves the practicality of real-time deployment in personal healthcare applications.

Taken together, these results indicate that stacking timing-based encoding, Mercer kernel representations, and explicit duration modelling leads to a cumulative performance gain. In particular, the proposed complete EvtTiSA model ({\color{color_C6}$\star$}) achieves an average AUC of 0.909, outperforming a standard Transformer encoder with sinusoidal order-based positional encoding ({\color{color_C3}$\blacktriangle$}) (0.883) as well as a variant without positional information ({\color{color_C4}$\boldsymbol{+}$}) (0.877). This comparison highlights the benefit of jointly modelling event timing and duration using adaptive temporal embeddings, beyond what can be achieved by conventional order-based positional encodings alone.

\subsection{Analysis of Event Contributions: Attention Attribution in Self-attention}

To analyse attention attributions, we averaged the integrated gradients across 165 trained models (15 random seeds of each of the 11 cross-validation folds). All analyses were conducted on the Mercer timestamps + duration models (EvtTiSA) with an input embedding dimension of 60, for which they achieved the best overall performance. For consistency analysis, each IG matrix was vectorised by concatenating its rows, and the 165 model instances were treated as independent random realisations of the same training procedure. ICC(1,1) and ICC(1,k) were computed for each sliding window to quantify single-model reliability and reliability of the averaged attribution, respectively. Averaging across the 63,113 sliding windows yielded a mean ICC(1,1) of 0.744 and a mean ICC(1,k) of 0.998, indicating substantial agreement among individual models and near-perfect reliability of the averaged attribution. We further observed that the baseline attention matrix $A'=\mathbf{0}$ yields $F(A')=F(\bm{0})=0.424$, which lies near the center of the output range $[0,1]$, suggesting that $A'=\bm{0}$ provides a reasonable neutral reference in Eq.~(\ref{eq_IG}).

Fig.~\ref{fig:ig_both} presents representative examples of attention attribution analysis for transition windows (a)--(c) and non-transition windows (d)--(f). Panels (a) and (d) show the attribution heatmaps, (b) and (e) show event durations, and (c) and (f) show event timestamps. In these heatmaps (a) and (d), each element $\mathrm{Attr}(A_{i,j})$ quantifies the contribution of the attention weight from event token $e_i$ to $e_j$ to the change in model output, $\Delta y$, indicated by color. Summing a row, $\sum_j \mathrm{Attr}(A_{i,j})$, aggregates the influence of $e_i$ on all other events, whereas summing a column, $\sum_i \mathrm{Attr}(A_{i,j})$, aggregates the influence of all events on the representation of $e_j$.


Across the dataset, column-wise attributions were more consistent than row-wise attributions. Column-wise analysis further revealed that long-duration head-central-movement event $h0$ often dominates the model decision when present. This observation is consistent with the manual annotation protocol, which relies on visible behavioural cues: annotated task transitions frequently co-occur with head movements (e.g., shifting from paperwork to laptop use typically involves raising the head to look at the screen). Such movements reduce the duration of the head-central state, thereby decreasing the prevalence of long-duration $h0$ segments around transitions. Fig. \ref{fig:ig_both}(a)--(c) illustrate a typical case, where the $h0$ column contributes strongly and negatively ($-0.423$) reducing the output to $0.001$. When $h0$ duration is shorter, as in Fig.~\ref{fig:ig_both}(d)--(f), other events such as $h+:head\_increase$ and $b+:is\_blink$ exhibit non-trivial contributions.

The attribution heatmaps also enable feature pruning based on aggregated column-wise contributions. We adopt column-wise aggregation not only due to its higher empirical consistency, but also because each column corresponds to a specific event token whose influence is further modulated by the value matrix $V$ that encodes the token content. Specifically, we summed column-wise attributions and aggregated them by event type to obtain the attribution contribution of each event type to $\Delta y$. These contributions were then averaged across sliding windows in which the corresponding event type occurred. The resulting rankings are reported in Table~\ref{tab:event-attribution-ranking}.

\begin{table}[htbp]
\caption{Average attribution per sliding window ($y=0.424+\Delta y\in[0,1]$), aggregated by event type.}
\begin{center}
\begin{tabular}{|c|c||c|c|}
\hline
\textbf{Event Type} & \textbf{Attribution} & \textbf{Event Type} & \textbf{Attribution} \\
\hline
$h0$ & $-0.348$ & $p0$ & $-0.051$ \\
$h+$ & $~~0.286$ & $p+$ & $~~0.050$ \\
$h-$ & $~~0.157$ & $p-$ & $~~0.024$ \\
$s-$ & $-0.112$ & $b-$ & $~~0.003$ \\
$b+$ & $~~0.102$ & $s+$ & $~~0.001$ \\
\hline
\end{tabular}
\label{tab:event-attribution-ranking}
\end{center}
\end{table}


\begin{figure}[pt]
\centerline{\includegraphics[width=\columnwidth]{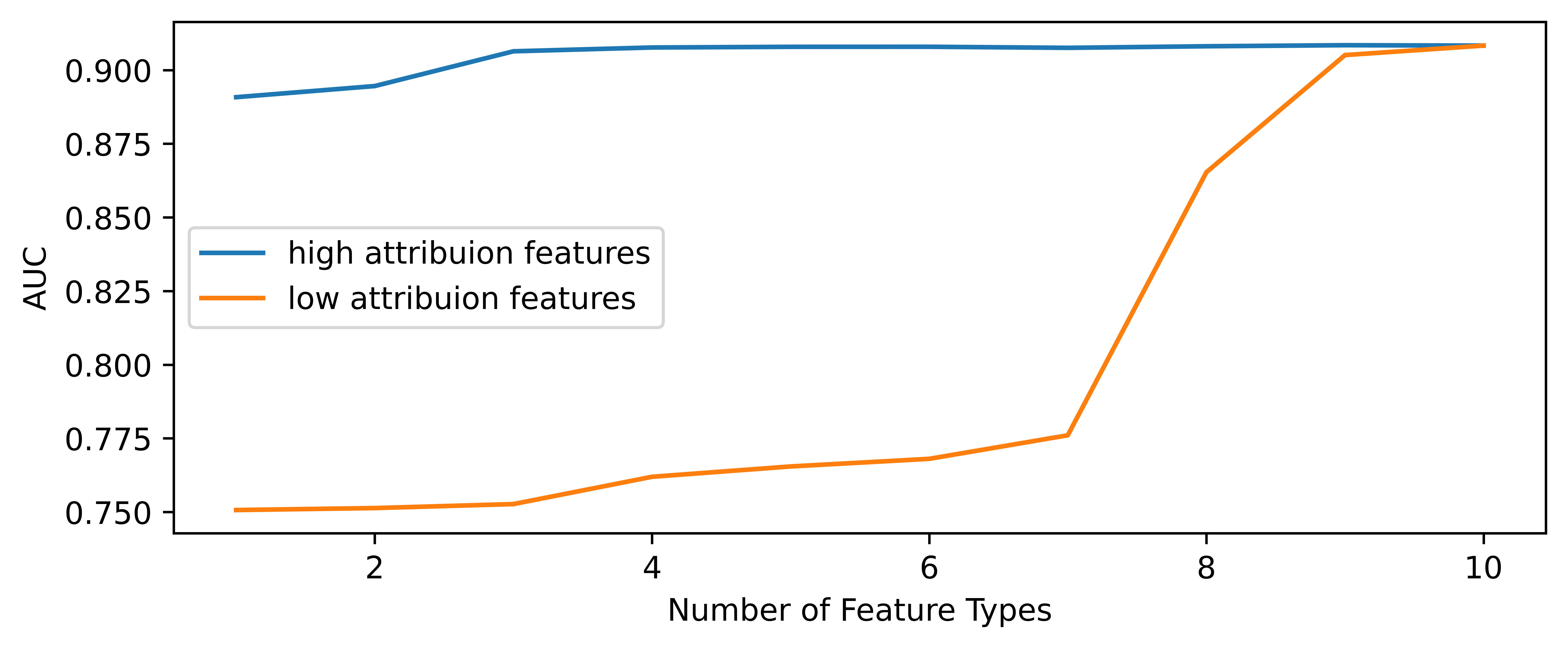}}
\caption{Effect of attribution-guided feature pruning on AUC performance.  Models were trained using either the most contributive event types (high-attribution features) or the least contributive event types (low-attribution features), as determined by Table~\ref{tab:event-attribution-ranking}. High-attribution features achieve strong performance with few event types, whereas low-attribution features require substantially more event types to approach comparable performance.}  
\label{fig:feature_pruning}
\end{figure}

\begin{table*}[htbp]
\caption{Performance comparison of different models on the task transition detection dataset. Values in parentheses indicate the standard deviation over 11 leave-one-participant-out cross validation folds. Among the cited models, only the SVM models were evaluated on this dataset in prior work~\cite{ouyang2024evaluation}; all other models were freshly implemented and trained under the same experimental protocol for fair comparison. Wilcoxon $p$-values were obtained from paired signed-rank tests comparing each model's AUC against EvtTiSA across the corresponding cross-validation folds.}

\begin{center}
\begin{tabular}{|c|c|c|c|c|c|}
\hline
\multirow{2}{*}{\textbf{Feature Type}} &
\multirow{2}{*}{\textbf{Model}} &
\multirow{2}{*}{\textbf{AUC}} &
\multirow{2}{*}{\textbf{Accuracy}} &
\multirow{2}{*}{\textbf{F1-Score}} &
\textbf{Wilcoxon test} \\
 & & & & & \textbf{$p$-value vs. EvtTiSA} \\
\hline

Statistical & SVM\cite{ouyang2024evaluation} & $0.911 (0.035)$ & $82.8\% (5.3\%)$& $72.1\% (7.1\%)$ & $5.20 \times 10^{-1}$\\
Deep & TST\cite{zerveas2021transformer} & $\textbf{0.918} (\textbf{0.032})$& $\textbf{83.1\%} (\textbf{4.8\%})$ & $\textbf{74.6\%} (\textbf{5.9\%})$ & $8.30\times10^{-2}$\\
Deep & LSTM & $0.915 (0.032)$ & $82.8\% (5.1\%)$& $73.9\% (4.9\%)$ & $6.74 \times 10^{-2}$\\

\hline

Event + Timing & SVM\cite{ouyang2024evaluation} & $0.888 (0.052)$ & $\textbf{82.8\%}~(\textbf{5.5\%})$ & $72.1\% (4.7\%)$ & $1.86\times 10^{-2}$ \\
Event + Order & Transformer Encoder 
\cite{vaswani2017attention} & $0.884 (0.039)$ & $79.0\%  (5.6\%)$ & $68.1\% (3.4\%)$ & $9.77 \times 10^{-4}$ \\
Event + Timing & TiSASRec\cite{li2020time}      & $0.894 (0.036)$ & $80.5\%  (5.0\%)$ & $70.6\% (3.4\%)$ & $9.77 \times 10^{-4}$ \\
Event + Timing & MEANTIME\cite{cho2020meantime} & $0.899 (0.034)$ & $81.0\%  (5.0\%)$ & $71.8\% (3.2\%)$ & $9.77 \times 10^{-4}$ \\
Event + Timing & \textbf{EvtTiSA*}              & $\textbf{0.909}~(\textbf{0.031})$ & $82.4\%~(4.6\%)$ & $\textbf{73.9\%}~(\textbf{3.5\%})$ & \textbf{-}\\

\hline
\end{tabular}
\label{tab:model_compare}
\end{center}
\end{table*}

According to Table~\ref{tab:event-attribution-ranking}, we conducted feature pruning using two complementary strategies: retaining event types with the highest absolute attributions, and retaining event types with the lowest absolute attributions. We then evaluated the resulting models to assess how attribution-based feature selection affects classification performance. As shown by the blue curve in Fig.~\ref{fig:feature_pruning}, which retains the highest-attribution features, a single-feature model using only $h0$ achieved an AUC of 0.891. Retaining the top three contributing event types, all of which are head-motion events, increased AUC to 0.906, indicating that head events alone provide strong predictive power for task transitions.

In contrast, a model retaining low-attribution features (orange curve), all of which are eye-related event types, achieved a substantially lower AUC of 0.776 despite including seven event types. Nevertheless, the most contributive eye events ($s-$ and $b+$) can partially substitute for head-motion events: a model using the feature set $\{h0, s-, b+\}$ achieved a comparable AUC of 0.900.

Beyond improving model transparency, the proposed attention attribution analysis provides a structured way to interpret behavioural information underlying task transitions. By identifying which event types dominate model decisions and validating their importance through attribution-guided feature pruning, this analysis supports a more principled understanding of how observable behaviours contribute to changes in human activity. Such insights are valuable for healthcare and health informatics applications, where they can inform sensor selection, reduce feature redundancy, and guide the design of efficient and interpretable monitoring systems for cognitive workload and functional-state assessment.







\subsection{Comparison of Models}

We compared the performance of the proposed EvtTiSA with various models using statistical, deep, and event-based features. As many of these models were originally evaluated on different datasets or in different research domains, we reproduced their implementations and trained them on our dataset following the descriptions in their respective papers. To the best of the authors' knowledge, self-attention–based event models have not previously been explored for human task processing; accordingly, this work implements and evaluates multiple Transformer-based event classifiers, including a standard encoder with order-based positional encoding, time-aware variants with timestamp encoding, and the proposed EvtTiSA model. 

For each cross-validation fold, performance was first averaged over 15 random seeds to reduce random initialisation variability; the reported mean and standard deviation were then computed across the 11 folds, and the resulting AUC-ROC, Accuracy, and F1-score are reported in Table~\ref{tab:model_compare}. SVM results were re-computed using the same 11-fold leave-one-participant-out cross-validation protocol, whereas the original publication~\cite{ouyang2024evaluation} reported concatenated-score evaluation. Given the substantial overlap in fold-level standard deviations, statistical significance was evaluated using paired Wilcoxon signed-rank tests on the corresponding fold-wise AUCs; the resulting $p$-values are reported in Table~\ref{tab:model_compare}.

As shown in Table~\ref{tab:model_compare}, the proposed event-and-timing-based EvtTiSA achieves the strongest overall performance among event-based neural models, outperforming Transformer encoders with order-based positional encoding as well as time-aware self-attention models originally developed for eCommerce applications, including MEANTIME~\cite{cho2020meantime} and TiSASRec~\cite{li2020time}. These improvements may be attributed to both the parameter-efficient Mercer kernel and the explicit modelling of event duration. Paired Wilcoxon signed-rank tests on the fold-wise AUCs confirm that these improvements over the event-based baselines are statistically significant ($p < 0.05$).

When compared with the event-based SVM model~\cite{ouyang2024evaluation}, EvtTiSA achieves comparable accuracy while providing consistent improvements in AUC-ROC and F1-score. On the one hand, this highlights the importance of incorporating event duration information in human task processing, as the SVM model also accounts for duration through handcrafted features. On the other hand, it suggests that self-attention–based models may require larger training datasets or effective transfer learning to achieve consistently dominant performance.

Nevertheless, we observe a clear improvement in AUC over the event-based SVM, which is statistically significant according to the paired Wilcoxon signed-rank test ($p = 0.0186$). This indicates that EvtTiSA exhibits improved discriminative ability, assigning higher confidence scores to transition instances than to non-transition instances more consistently across decision thresholds. Such an improvement is especially relevant in the context of wearable task transition detection, where operating points may vary across applications and users, and different scenarios may prioritise precision or recall differently. A higher AUC therefore enables more flexible threshold selection at deployment time, improving robustness under varying class imbalance and practical usage conditions.

From an interpretability perspective, the SVM model relies on bag-of-\emph{N}-gram features, which aggregate event patterns and make it difficult to attribute model decisions to individual events. In contrast, while neural-network-based models allow gradient-based attribution, they often lack a well-defined static neutral reference state, which complicates the faithful application of integrated gradients. By operating on explicit event tokens with well-defined temporal semantics, EvtTiSA provides a more interpretable representation that better supports event-level attribution analysis.



For non-event-based approaches, we considered three representative reference systems: (1) a statistical-feature-based SVM~\cite{ouyang2024evaluation}, (2) the Time Series Transformer (TST)~\cite{zerveas2021transformer}, a state-of-the-art attention-based model for time-series processing, and (3) an end-to-end LSTM. As shown in Table~\ref{tab:model_compare}, statistical-feature-based and deep time-series models achieve performance comparable to that of EvtTiSA. In particular, the difference in AUC between EvtTiSA and the statistical SVM is not statistically significant according to the paired Wilcoxon signed-rank test ($p = 0.52$), indicating comparable discriminative performance. Similarly, EvtTiSA and TST exhibit very similar results across evaluation metrics; although TST achieves a slightly higher mean AUC, the difference is not statistically significant ($p > 0.05$).

Despite using a substantially more compact event-based representation, EvtTiSA achieves performance on par with deep models operating directly on raw time series. This indicates that event-based representations augmented with timing information preserve most of the discriminative information contained in the original signals. Moreover, while deep-feature-based models such as TST and LSTM rely on latent representations that are difficult to interpret, EvtTiSA benefits from more transparent feature semantics and enables event-level attribution analysis, as discussed in Section~V-B, without sacrificing performance.

It is worth noting that we also implemented several high-performing models previously proposed for human task transition detection, including a series-connected CNN–LSTM~\cite{wang2020wearable} and an ensemble CNN–LSTM~\cite{khan2025transition}. In our experiments, however, these models exhibited strong overfitting on the present dataset, yielding AUC, accuracy, and F1 scores of approximately 80\%, 73\%, and 40\% respectively, and are therefore not included in Table~\ref{tab:model_compare}. This behaviour is likely attributable to the large number of parameters introduced by the multilayer CNN components relative to the dataset size. Notably, the performance gains of these deep architectures over the SVM trained with event-based statistical features were similarly modest, further supporting the observation that deep learning models for human task processing may require larger datasets or effective transfer learning strategies to substantially outperform conventional approaches.

Despite these encouraging results, several limitations should be noted. First, the dataset comprises recordings from 11 participants in a semi-naturalistic laboratory setting, which may limit the generalisability of the findings to larger and more diverse populations. Second, although class imbalance was mitigated through transition-window oversampling, real-world deployment may involve different imbalance ratios and activity distributions. Evaluating the framework on larger-scale and fully in-the-wild datasets will therefore be an important direction for future research.

\section{CONCLUSION}

This work presented EvtTiSA, an event-based, time-aware self-attention framework for wearable task transition detection. By representing behavioural data as compact temporal event tuples and explicitly modelling both event onset timing and event duration, the proposed approach addresses key limitations of conventional self-attention models that rely solely on token order. The use of a Mercer kernel enables flexible and parameter-efficient timestamp encoding, while the proposed type-aware duration embedding allows persistent events to influence attention in a controlled and interpretable manner.

Across extensive cross-validated experiments, EvtTiSA achieved improved or comparable performance to representative statistical, deep, and event-based models, with strong discrimination as measured by the area under the receiver operating characteristic curve. Importantly, these results were obtained using a substantially more compact representation than uniformly sampled event indicators or raw time-series inputs, highlighting the efficiency advantages of temporal event modelling for wearable systems with limited computational resources.

Beyond performance, this work emphasised interpretability as a priority design objective. By enforcing a well-defined neutral reference state for attention, integrated-gradient-based attribution analysis was shown to yield stable and meaningful event-level explanations. The resulting attribution patterns aligned with known behavioural cues of task transitions and enabled effective attribution-guided feature pruning, supporting principled sensor and feature selection.

Overall, the findings demonstrate that incorporating explicit temporal structure---particularly event duration---into self-attention models enhances both predictive capability and transparency in human task modelling. While the current study focuses on task transition detection, the proposed framework is broadly applicable to other event-driven wearable physiological and behavioural analysis tasks. Future work will explore larger-scale datasets, cross-domain transfer learning, and pre-training strategies to further improve generalisability and unlock the full potential of time-aware event-based attention models.

\section*{Acknowledgment}

This work was supported in part by US Army Cooperative Agreement
W911NF1920330. Opinions expressed are the authors’, and may not reflect
those of the US Army.

\section*{References}

\def\refname{\vadjust{\vspace*{-2.5em}}} 

\bibliographystyle{IEEEtran}
\bibliography{ref.bib}

\end{document}